\documentclass[journal]{IEEEtran}
\usepackage{color}
\usepackage{colortbl}
\usepackage[colorlinks,linkcolor=red,anchorcolor=blue,citecolor=green]{hyperref}
\usepackage{cite}
\usepackage{soul}
\ifCLASSINFOpdf
\else
\fi

\usepackage{graphicx}
\usepackage{color}

\begin{document}
%




\title{Landslide4Sense: Reference Benchmark Data and Deep Learning Models for Landslide Detection}

\author{Omid Ghorbanzadeh, 
        Yonghao Xu,~\IEEEmembership{Member,~IEEE},  
        Pedram Ghamisi,~\IEEEmembership{Senior~Member,~IEEE}, 
        \\Michael Kopp, and 
        David Kreil
\thanks{O. Ghorbanzadeh, Y. Xu, P. Ghamisi, M. Kopp, D. Kreil are with the Institute of Advanced Research
in Artificial Intelligence (IARAI), 1030 Vienna, Austria (email: omid.ghorbanzadeh@iarai.ac.at; yonghao.xu@iarai.ac.at; pedram.ghamisi@iarai.ac.at; michael.kopp@iarai.ac.at; 	
david.kreil@iarai.ac.at).}
\thanks{P. Ghamisi is also with the Helmholtz-Zentrum Dresden-Rossendorf, Helmholtz Institute Freiberg for Resource Technology, Freiberg 09599, Germany.}
\thanks{Manuscript received in 2022. Corresponding authors: Yonghao Xu; Omid Ghorbanzadeh.}}

\markboth{IEEE Transactions on Geoscience and Remote Sensing,~Vol.~60, October~2022}%
{Shell \MakeLowercase{\textit{et al.}}: Bare Demo of IEEEtran.cls for IEEE Journals}
%

\maketitle

\begin{abstract}
This study introduces \textit{Landslide4Sense}, a reference benchmark for landslide detection from remote sensing. The repository features 3,799 image patches fusing optical layers from Sentinel-2 sensors with the digital elevation model and slope layer derived from ALOS PALSAR. The added topographical information facilitates an accurate detection of landslide borders, which recent researches have shown to be challenging using optical data alone. The extensive data set supports deep learning (DL) studies in landslide detection and the development and validation of methods for the systematic update of landslide inventories. The benchmark data set has been collected at four different times and geographical locations:  Iburi (September 2018), Kodagu (August 2018), Gorkha (April 2015), and Taiwan (August 2009). Each image pixel is labelled as belonging to a landslide or not, incorporating various sources and thorough manual annotation. We then evaluate the landslide detection performance of 11 state-of-the-art DL segmentation models: U-Net, ResU-Net, PSPNet, ContextNet, DeepLab-v2, DeepLab-v3+, FCN-8s, LinkNet, FRRN-A, FRRN-B, and SQNet. All models were trained from scratch on patches from one quarter of each study area and tested on independent patches from the other three quarters. Our experiments demonstrate that ResU-Net outperformed the other models for the landslide detection task. We make the multi-source landslide benchmark data (Landslide4Sense) and the tested DL models publicly available at \url{https://www.iarai.ac.at/landslide4sense}, establishing an important resource for remote sensing, computer vision, and machine learning communities in studies of image classification in general and applications to landslide detection in particular.
\end{abstract}


%
\IEEEpeerreviewmaketitle

\section{Introduction}
%
%
%
%

\IEEEPARstart{L}{andslides} are a natural hazard commonly found in mountainous terrain on all continents and posing a significant concern in these regions \cite{Guzzetti2006}. Strong earthquakes, exceptional meteorological events like intense rainfalls, volcanic activity, and anthropogenic activities such as constructing road networks that crossed the slopes are the primary landslide triggers, posing high risks to properties and society \cite{Lima2017}. The first two of these triggers are by far the leading causes of numerous small to large-sized destructive landslides over large areas \cite{Azmoon2021,Prakash2021}. The frequency of this natural hazard is increasing due to current climate changes, posing a significant threat to sustainable development in mountainous areas \cite{Ghorbanzadeh2021}. For example, the seismic shaking and aftershocks of an intense earthquake with a magnitude (Mw) of 6.56 that happened on September 6, 2018, exactly one day after the powerful Typhoon Jebi, triggered over 5600 landslides distributed in an area of 46.3 km$^2$ in Eastern Iburi, Hokkaido, Japan \cite{Zhang2019c}. In a report from the Disaster Risk Reduction Office of the United Nations, landslides and mass movements account for 5.4\% of all climate-related disasters that have occurred worldwide over the past two decades. Of all the reported disasters from 1998 to 2017, landslides and mass movements affected about five million people \cite{Wallemacq2018}. The considerable population size affected is mainly due to current rapid urbanization and population growth in landslide-prone regions \cite{Guzzetti2022}. Although most of the landslides that take place annually are categorized as small, these are the large ones that are multiplied by the consequences and are responsible for most of the damages and casualties \cite{dai2020entering}. Therefore, rapidly monitoring the spatial distribution and location of landslides and recording them in a centralized landslide inventory dataset is essential for estimating the damages they impose \cite{Guzzetti2012}. If collected promptly, this critical information about the ground surface of an affected area enables decision-makers to evaluate the impact of landslide consequences on human vulnerability criteria, including socioeconomic and environmental implications \cite{pradhan2015data}. In addition to promptly evaluating the risk and disaster responses, the landslide inventory dataset is a fundamental prerequisite for advancing the understanding of this complex phenomenon. Although there is no standard for a landslide inventory dataset or archive \cite{Galli2008}, the best cases are those that are post-processed and organized based on definitions for example, those provided by Cruden and Varnes 1996 \cite{Cruden1996} related to categorization and distribution of activity. However, the primary data required are the landslide's exact extent, shape, length, size, number, and border, which can be used to extract advanced information such as whether it is shallow or deep-seated and even its kinematic mechanisms. The other important condition of such a landslide inventory dataset is need to update the inventories multi-temporally to get precise information about landslide triggering period, lifetime, and velocity \cite{Guzzetti2006}. Since this phenomenon does not happen randomly, a landslide inventory can contribute to disaster preparedness and support early warning systems to warn government leaders and the population at risk \cite{Yang2022, noviello2020monitoring}. This can be done by using spatio-temporal variabilities from the landslide inventory dataset and related thematic environmental factors to model and map the geographic location and probability of future landslides based on the same conditions and triggers that caused previous landslides \cite{Corominas2014,Gariano2016,Khaing2020}. Moreover, the resulting landslide probability maps determine how prone a landscape is to cause landslides, which is essential for landscape management and preventing land degradation dynamics and consequences \cite{Titti2021}. 

A landslide inventory dataset can be generated using different methods \cite{Galli2008}. Although a field survey is deemed one of the most reliable ways o create and update landslide inventory datasets, it is usually time-consuming and dangerous for large study areas and remote regions, respectively \cite{ghorbanzadeh2019evaluation}. Also, manually deleting landslides through visual interpretation methods based on aerial imagery are time- and resource-intensive, requiring skilled experts for assembly and updating \cite{Guzzetti2012}. 

Earth observation data and satellite imagery are considered the most available sources of the data for landslide detection, especially in large areas \cite{Shi2020}. Although this geologic phenomenon leaves discernible appearances and signatures, landslide detection from satellite imagery—and hence updating the inventory dataset—is a challenging task due to the low spectral heterogeneity between the existing surface features and landslides in the image; the diverse climate categories, geology, and geomorphology characteristics in different landslide-affected areas also present challenges. Moreover, since a landslide is usually the secondary disaster that takes place during or after a trigger phenomena, its characteristics are significantly related to the trigger. Semi-automated landslide detection approaches have been developed based on machine learning models, some exploiting satellite imagery but mainly making use of images from commercial satellites or unmanned aerial vehicles \cite{Karantanellis2021}. These approaches consist of two main phases: using pixel or object-based methods to extract the landslide's representative features, and then applying a supervised machine learning model such as support vector machine (SVM) and random forest (RF) models to classify the features as landslide and non-landslide pixels or segments \cite{TavakkoliPiralilou2019}. The object-based semi-automated landslide detection approaches usually outperformed the pixel-based ones \cite{Keyport2018}. Since the former method uses homogeneous image objects, it can take to account the textural and morphological characteristics of landslides in addition to merely the radiometric characteristics from every single pixel \cite{Blaschke2010}. Also, including expert knowledge and user-defined parameters for determining the optimum parameters, thresholds for landslide segments in object-based methods help to provide higher accuracy than pixel-based methods \cite{Pawuszek2019}. However, current object-based methods often address mainly the low-level features of individual objects, which limits their generalization potential, since there is also a need to include high-level features such as object topologies \cite{Ghorbanzadeh2022}. Additionally, although creating handcrafted high-level features using expert knowledge is transparent and flexible compared to deep learning (DL) models, it is laborious, time-consuming, and subjective \cite{Amatya2021}. 

The advent of the era of DL models like deep convolutional neural networks (DCNN) has provided unique opportunities for landslide detection from satellite imagery, mainly on a large scale \cite{Ghorbanzadeh2021,gawlikowski2022advanced}. Thus far, these models have usually followed a standard supervised learning framework based on satellite imagery of landslide-affected areas from different geographical locations and corresponding labeled inventories for training and testing the models \cite{Prakash2021}.  
The main contributions of this paper are as follows:
\begin{enumerate}
  \item We examine the usage of freely accessible satellite data collected from four geographical regions across the globe that have been affected by landslides due to different trigger.
  \item We introduce a multi-source landslide benchmark dataset consisting of 3,799 image patches by fusing optical layers from Sentinel-2 with the DEM and slope layers derived from ALOS PALSAR.
  \item We evaluate the landslide detection performance of 11 state-of-the-art DL segmentation models based on the provided benchmark data set.
  \item We provide a publicly available landslide benchmark and share our research findings to promote current and future advancements in machine learning and computer vision within this field.
\end{enumerate}
The presented study is organized as follows. Section II briefly reviews the current state-of-the-art DL models for landslide detection and surveys the applied satellite imagery. In Section III, we then describe our landslide benchmark data (Landslide4Sense) and its division into the training and testing segments. We concisely introduce selected state-of-the-art DL segmentation models and implement them for the training segment of the Landslide4Sense benchmark data in Section IV. Finally, we use the testing segment to evaluate the models' generalization performance and discuss the segmentation results in Section V. Our conclusions are presented in Section VI.

\section{Prior work}
Annotated datasets have become increasingly important in the era of DL. There have been numerous datasets including different data types published over the last decade, each designed for a particular remote sensing application such as semantic segmentation (e.g., land cover classification \cite{wang2021loveda}), instance segmentation (e.g., Vehicle footprint extraction \cite{mou2018vehicle}), and object detection and localization (e.g., Airplane detection \cite{wang2022remote}). Therefore, the availability of such datasets to train and validate deep learning algorithms becomes a prerequisite for advanced research in remote sensing \cite{budde2022development}. Depending on the purpose of the project, researchers may use RS imagery acquired from one source or a dataset compiled from RS imagery from multiple sources. Additionally, given the current volume and velocity of RS imagery, they differ significantly in terms of some critical parameters such as spatial resolution, temporal resolution, price, and availability.
To the best of the authors' knowledge, there is not yet a multi-source dataset to provide globally distributed case study areas of landslide-experienced regions for landslide detection from freely available RS imagery. In most cases, the landslide dataset development activities have been carried out on an event-by-event basis at the local level. A good example can be found in Nepal's prepared wide range of landslide inventories following the Mw 7.8 (Gorkha) and Mw 7.3 (Dolakha) earthquakes on 25 April 2015. The landslide inventories were made in two main formats, polygonal and point-based, and the number of landslides for the same affected area of 10,000 km$^2$ varied. In one of the most comprehensive studies, applying various methodologies including visual interpretation, \cite{roback2018size} created an inventory of almost 25,000 landslides with a total area of 87 km$^2$ from high-resolution satellite imagery (Worldview-2 and -3 with spatial resolution of 20–50 cm) acquired mainly from 2 to 8 of May 2015. In another study, \cite{valagussa2016regional} generated three landslide inventory data sets for an area that includes four districts in the North of Kathmandu, Nepal's capital. Individual polygons represent each landslide. Based on multi-temporal images consisting of Bing maps, Google Crisis maps, and Google Earth, as well as videos taken by helicopter, the first data set reported landslides that occurred immediately following the earthquake and the aftershocks. More than 15,500 landslide polygons covering a total area of over 90 km$^2$ were detected by \cite{martha2017spatial} for the same event. Similar to \cite{roback2018size}, they also did visual image interpretation using VHR satellite images such as Cartosat-1 and 2, WorldView, Resourcesat-2, and the Pleiades with spatial resolutions ranging from 0.3 m to 5 m.
The number of DL studies involving landslide detection based on satellite imagery has sky rocketed since 2019, and several DL networks and the leading CNN model \cite{Krizhevsky2012} have been developed for landslide detection from a wide range of satellite imagery and remote sensing data. Here we summarize the common DL models and data that have been widely applied for landslide mapping. The studies in this field were pioneered by Ghorbanzadeh \emph{et al.} \cite{ghorbanzadeh2019evaluation}, and Ye \emph{et al.} \cite{Ye2019}, who proposed different DL models architectures for mapping landslides from very high resolution (VHR) satellite imagery and hyperspectral images, respectively. In \cite{ghorbanzadeh2019evaluation}, two proposed CNN architectures and three machine learning models of SVM, RF, and artificial neural networks were trained from scratch using RapidEye VHR satellite imagery (5 m spatial resolution) and topographic factors like slope angle information. The trained models were then used for landslide detection in a part of Nepal that did not apply to the training phase. For the most part, the proposed CNNs outperformed the machine learning models. YOLOv5 was applied by Wang \emph{et al.}  \cite{Wang2021} for landslide detection based on high-resolution remote sensing images. They were able to improve its performance and increase its detection accuracy by 1.64\% using an attention module.  Yu \emph{et al.} \cite{Yu2021} introduced the Matrix SegNet for landslide mapping from satellite imageries with different spatial resolutions. The DL models of DenseNet and CNN have been compared by \cite{Liu2021} for landslide inventory construction for their study area in Hubei Province, China. Their calculated Kappa coefficient was 0.965 for DenseNet, and 0.908 for the applied CNN model. They have proposed a landslide detection model of a contour-based semantic segmentation developed based on the pyramid scene parsing network (PSPNet), which was applied for the annual Landsat image acquired in Nepal. Their F1 score was 60\% for Nepal's national case landslide detection. A deep CNN model, named LanDCNN, was developed by Su \emph{et al.} \cite{Su2021} for landslides segmentation in Lantau Island. The model employs VHR bitemporal RGB aerial images and a digital terrain model with a resolution of 0.5 m acquired by the GEO of Hong Kong.

Applications of the fully convolutional network (FCN) \cite{Long2015} and U-Net \cite{Ronneberger2015} for landslide detection have been demonstrated by \cite{Lei2019} and built upon by many studies using different variants of these models. For instance, the U-Net and ResU-Net models have been compared by Gao \emph{et al.} \cite{Gao2021} and Ghorbanzadeh \emph{et al.}  \cite{Ghorbanzadeh2021} for mapping landslides using Landsat-8 OLI and Sentinel-2 images, respectively. The latter study has performed 48 different scenarios to evaluate the generalization and transferability of these models for landslide detection in case studies in three different areas of Eastern Iburi, Shuzheng Valley, and Western Taitung County. They find that ResU-Net showed higher generalization through the use of the training and testing data from areas with similar characteristics, but the U-Net model demonstrates slightly better transferability among different geographical regions. These models have been applied by Qi \emph{et al.} \cite{Qi2020} to attempt to detect the landslides triggered by heavy rainfall in Tianshui City, Gansu Province in July 2013. The resulting F1 scores for ResU-Net and U-Net were 0.89 and 0.8, respectively obtained by training and testing based on 3-bands of near-infrared, red, and green VHR GeoEye-1 images with a spatial resolution of 0.5 m. Yi \& Zhang \cite{Yi2020} proposed a network that could outperform both the ResU-Net and U-Net for landslide mapping from RapidEye images. Yokoya \emph{et al.} \cite{Yokoya2020} applied U-Net and LinkNet models for  mapping of debris flow using remote sensing optical imagery and topographic data. Lv \emph{et al.} \cite{Lv2020} and Ghorbanzadeh \emph{et al.} \cite{Ghorbanzadeh2022a} applied multi-temporal images from the pre- and post-event for landslide inventory mapping using a dual-path FCN and an object-based image analysis (OBIA)-ResU-Net model, respectively. A landslide detection deep network was designed by Zhang \emph{et al.} \cite{Zhang2020b} based on a two-stream U-Net with shared weights to discover the relevant deep features from bi-temporal images. Liu \emph{et al.} \cite{Liu2020a} modified the standard U-Net to provide an automatic landslide identification model implemented for images with 0.14 m and 0.47 m spatial resolutions for testing and training processes, respectively. Soares \emph{et al.} \cite{Soares2020} evaluated the impact of data augmentation, adding elevation information, locations, and sizes of sample patches, on the performance of the U-Net model from landslide detection from RapidEye images. The impact of the topographical information of elevation, slope, aspect, and plan curvature on the overall performance of a multi-stream CNN-based landslide detection is assessed using the Dempster–Shafer model in \cite{Ghorbanzadeh2020a}. Not only optical images but also synthetic aperture radar (SAR) data have been applied for landslide detection using DL models, employing totally different imaging geometries and content \cite{Zhu2017}.  Nava \emph{et al.} \cite{Nava2021} applied SAR data derived from Sentinel-1 for training and testing a CNN model, rather than using Sentinel-2 images. However, Machielse \emph{et al.} \cite{Machielse2021}  applied DL models of U-Net and a Generative Adversarial Network (GAN) to map landslides from SAR data, and both models failed. They recommend considering a longer time for temporal averaging of SAR data in order to obtain acceptable outcomes, but this approach is impractical for emergency cases and timely landslide mapping. The use of SAR data is mainly applicable for cases in which there is no access to cloud-free images after the event, which is very common for rainfall-induced landslides. The DL model of GAN was also used by \cite{Fang2020a} in a Siamese neural network for landslide inventory mapping based on VHR images (0.5 m spatial resolution) acquired in the west of Lantau Island.
Unmanned aerial vehicle (UAV) data was also the primary data source for landslide mapping using different DL models like CNNs \cite{Catani2021,Ghorbanzadeh2019c}, and Mask R-CNN \cite{Ullo2021}. In \cite{Ghorbanzadeh2019c}, a quadcopter UAV (DJI Mavic 2 Pro) was used for image acquisition from two case study areas in the higher Himalayas, and they applied different CNN models for landslide detection in the UAV images. Their applied approach got the highest F1 score of 85\% and a mean intersection-over-union (mIOU) of over 74\% using sample patches with a size of 64 $\times$ 64 pixels. 

In summary, in the wake of recent progress in computer vision technologies, computational resources, and satellite imagery availability, DL models are taking off in landslide detection, as they have in other remote sensing fields. Our literature review also indicates that the application of DL models for landslide detection is one of the fastest-growing new approaches and is proving to be very successful in this field. However, the application of DL models for landslide detection from remote sensing data raises some new challenges, such as these models' transferability capabilities when confronting novel geographical areas with different landcover and morphological characteristics. To the best of our knowledge, most of the implemented and developed DL models have been evaluated in a local geographical region, typically covering a small area divided into training and testing areas using different ratios \cite{Ghorbanzadeh2021}. Therefore, the direct applicability of these models to a novel unexplored geographical region, mainly in emergency cases, is usually unclear \cite{Prakash2021,Gao2021}. 
Also, DL models are adequately trained when an extensive training dataset with annotated landslides is used to learn effective models with several different parameters. The lack of publicly available datasets of this type is considered an obstacle that prevents the extensive use of DL models in landslide detection. Furthermore, the models have been created based on multimodal satellite imageries such as optical (multi- and hyperspectral) and SAR data. A trained DL model with data from a specific sensor is not usually practical for the others. Thus, there is no explicit norm to compare the generalization and transferability abilities of several DL models proposed for landslide detection during the past three years. In order to address such challenges and promote a new trend of promising research, we provide a landslide benchmark archive based on freely available satellite data collected from four diverse geographical regions; we then evaluate 11 fruitful DL models that were recently offered for computer vision tasks.
\begin{figure*}[t]
\centering
\includegraphics[width=0.9\textwidth]{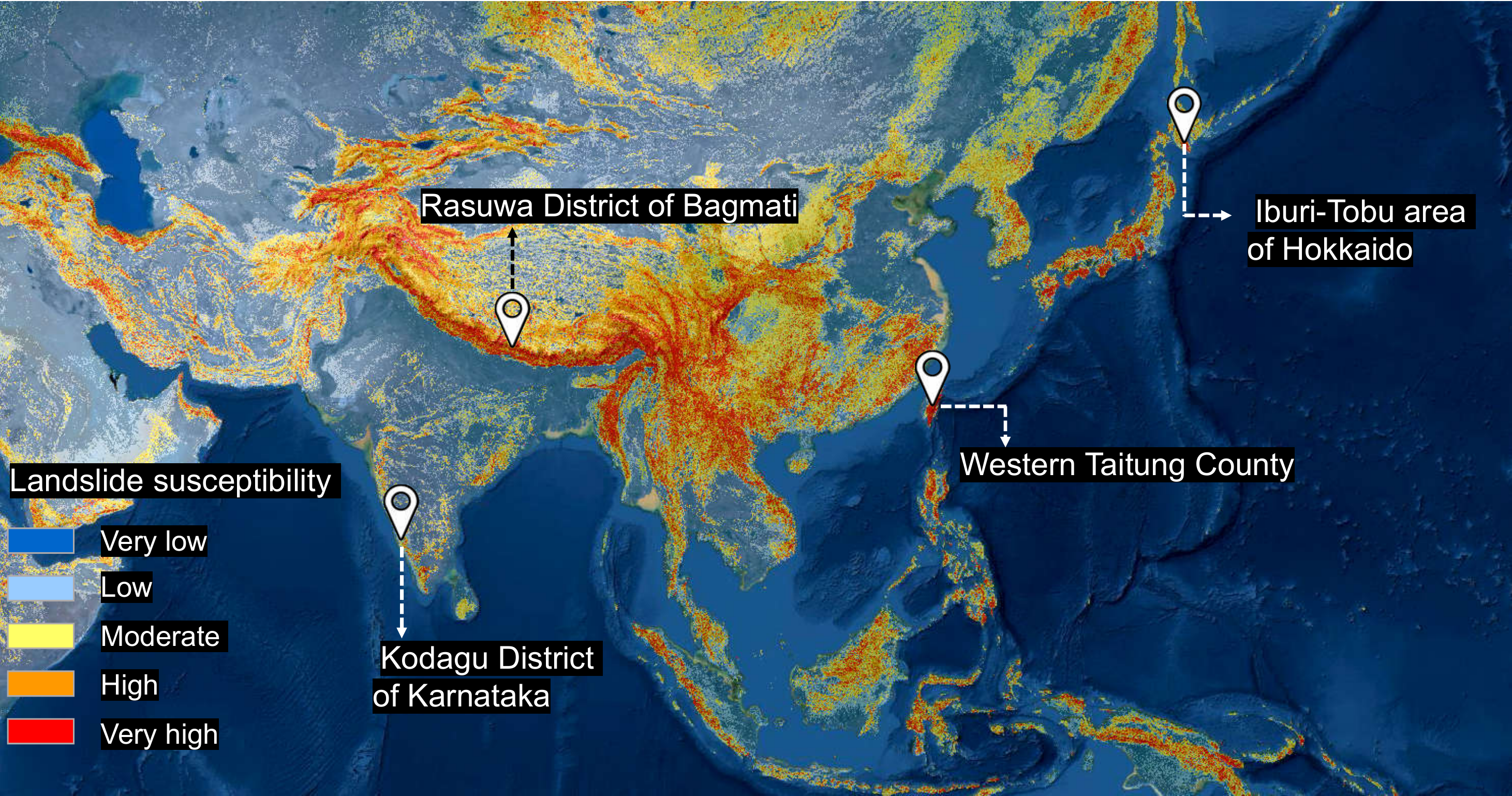}
\caption{Representing the Geo-locations of the selected case study areas on a global landslide susceptibility map.}
\label{Fig: Landslide susceptibility map}
\end{figure*}

\section{Dataset description}
This study aims to devise a multi-source landslide benchmark dataset that addresses the need to annotate data in order to train DL models for the landslide detection task. Although it is not clear how much data is required to get the best performance from supervised DL models, it has been proven that a small training dataset of labeled images results in poor classification. Moreover, any big training dataset, even one including enormous quantities of labeled data, may not include all contingent situations \cite{kitchin2016makes}. Therefore, a DL model that is trained based on a dataset (no matter how large) may yield a limited performance when it encounters a novel case not observed during the training \cite{marcus2018deep} or out-of-distribution input \cite{alcorn2019strike}. This matter is prevalent in the case of landslides, which usually exhibit a wide range of sizes, shapes, and geographical locations with different characteristics. Thus, we intentionally have selected landslide-experienced study areas from four different geographical regions to increase the volume as well as the variety in the Landslide4Sense benchmark data. In addition, the landslides in these study areas were initiated by various triggers, sometimes in combination, which can enhance the transferability capability of the DL models trained by this benchmark data to other regions. In this section we introduce a) our selected case study areas, b) the characteristics of Sentinel-2 and ALOS PALSAR sensors, c) the way of landslide inventory annotation, d) the statistics and shape of the dataset.

\subsection{Study Areas}
The exact geographic locations of the four different study areas are represented over a background of a global landslide susceptibility map generated by \cite{stanley2017heuristic} (see Figure \ref{Fig: Landslide susceptibility map}). This map is generated by a heuristic fuzzy approach using five explanatory variables: slope degree (generated mainly from Shuttle Radar Topography Mission (SRTM) \href{https://www2.jpl.nasa.gov/srtm/}{https://www2.jpl.nasa.gov/srtm/}), forest loss (Landsat-based global map produced from 2000 to 2013 by \cite{hansen2008humid}), geology (Geological Map of the World (GMW) \cite{bouysse2009geological}), road networks (OpenStreetMap (OSM) \href{https://www.openstreetmap.org/}{https://www.openstreetmap.org/}), and faults (GMW).

\begin{figure}
\includegraphics[width=0.5\textwidth]{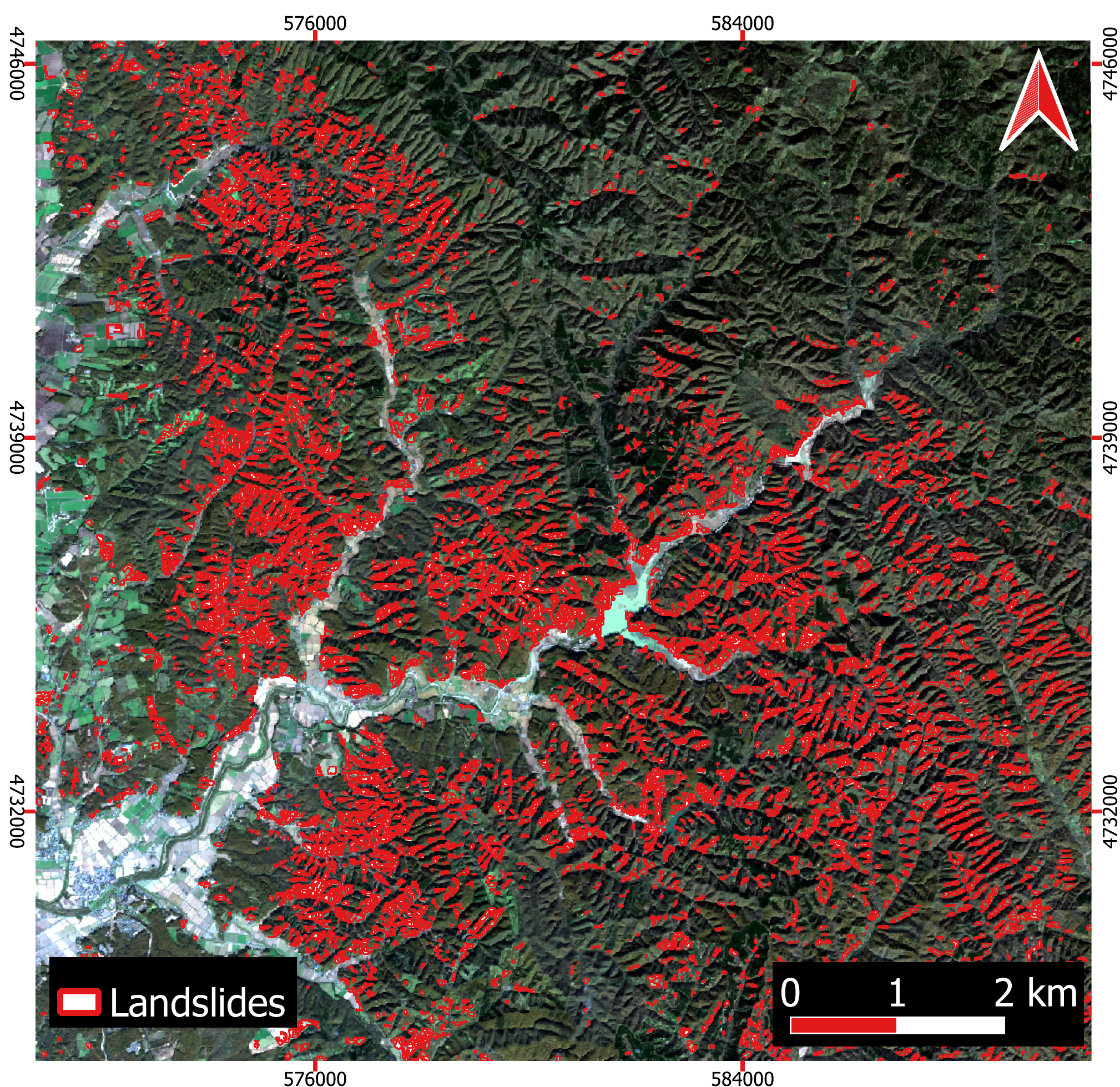}
\caption{Iburi-Tobu area of Hokkaido.}
\label{Fig: Iburi-Tobu area of Hokkaido}
\end{figure}
\subsubsection{Iburi-Tobu Area of Hokkaido}
The Iburi-Tobu area is located in the southern part of the second-largest island of Japan, named Hokkaido. This area was hit by an earthquake with a magnitude (Mw) of 6.6, a maximum intensity of 7, and a series of aftershocks on September 6, 2018. It caused many secondary geo-disasters like large-scale shallow-sliding landslides distributed over hilly regions, valley damming by landslides, and quake lakes formed in the streams \cite{Yamagishi2018}. Extensive damages to public infrastructure was also reported, including destruction of roads, bridges, and power stations, cut power lines, and an increase of the turbidity in reservoir dams as more sediment flowed into streams after the landslides \cite{Chen2001}. Over 5600 landslides occurred in the Iburi-Tobu region, due to an earthquake that occurred after three days of continuous rainfall due to typhoon Jebi, which accumulated close to 100 mm precipitation  \cite{Zhang2019c}. A number of landslide inventories have been generated for the Iburi-Tobu region since the earthquake, for example by the Geographical Survey Institute (GSI) of Japan \cite{ishikawa2021reconnaissance} and Zhang \emph{et al.} \cite{Zhang2019c}, using VHR aerial images. 

\begin{figure}
\includegraphics[width=0.5\textwidth]{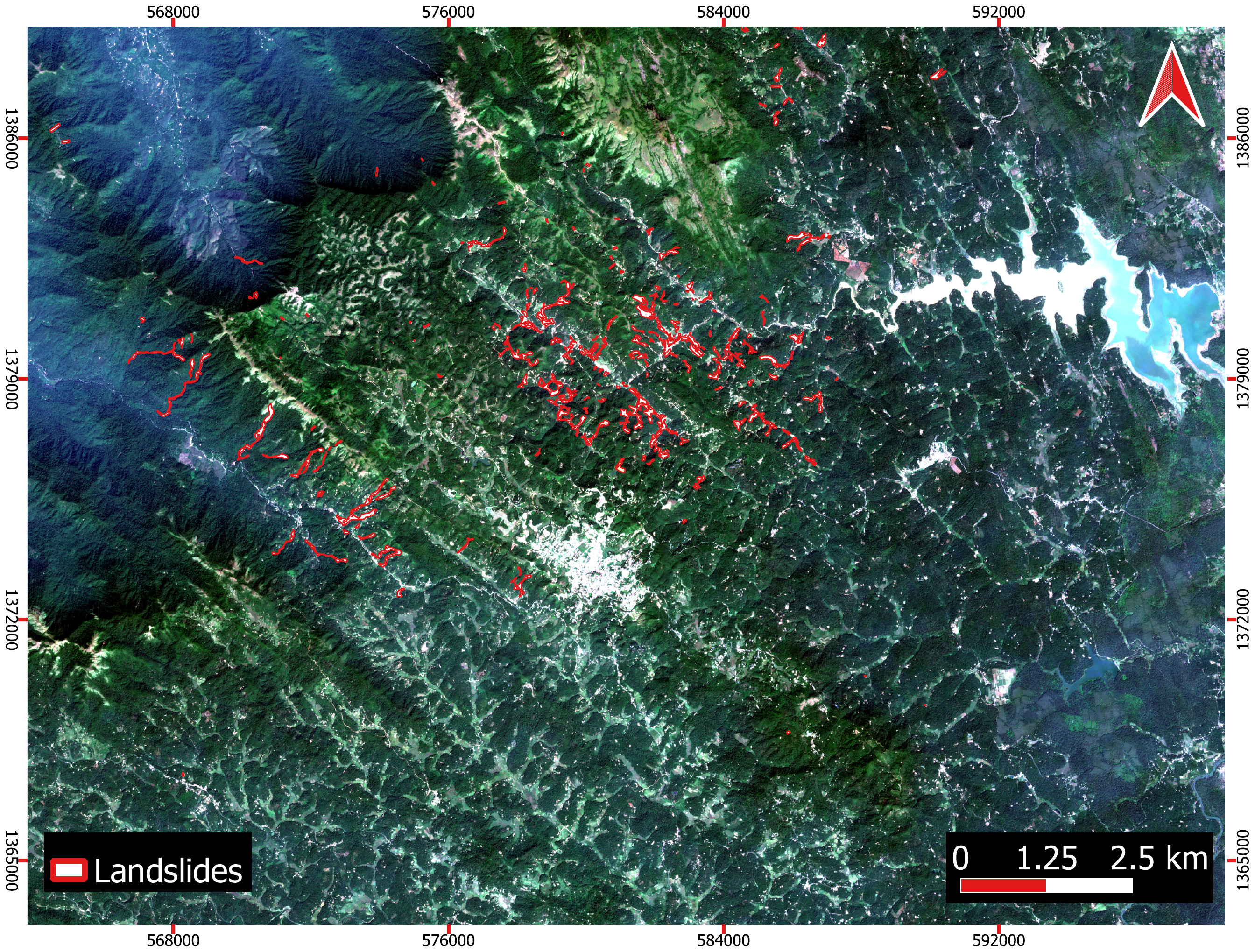}
\caption{Kodagu District of Karnataka.}
\label{Fig: Kodagu District of Karnataka}
\end{figure}

\subsubsection{Kodagu District of Karnataka}
In August of the late monsoon season of 2018, exceptional rainfall occurred in most parts of India, including the Kodagu District of Karnataka, due to current changing precipitation patterns in this region. Over one month, the total rainfall of more than 1200 mm triggered severe landslides and flash floods in this rural district. With this rainfall, several landslides occurred, many of them were debris flows, claiming 16 lives and damaged around 200 villages \cite{Shahabi2021}. In this study area, the dominant geomorphological setting consists of highly dissected and sloping structural hill ranges. The land cover of the Kodagu district is mainly occupied by rice and coffee agricultural fields through cultivated valleys and agroforestry crops such as cardamom and pepper in dense forests. However, over the past two decades this area has also become known for wide-scale land cover and land-use changes such as unplanned urbanizations, deforestation, and mining. Therefore, the disturbances of these anthropogenic criteria and the altered precipitation patterns have significantly aggravated rainfall-triggered landslides in this region \cite{yunus2021unraveling}. The whole or some parts of landslide-affected areas in this district have already been evaluated for landslide detection by different researchers. For example, Shahabi \emph{et al.} \cite{Shahabi2021}  developed an unsupervised learning approach to detect landslides in the Kodagu district without using any inventory dataset.

\begin{figure}
\includegraphics[width=0.5\textwidth]{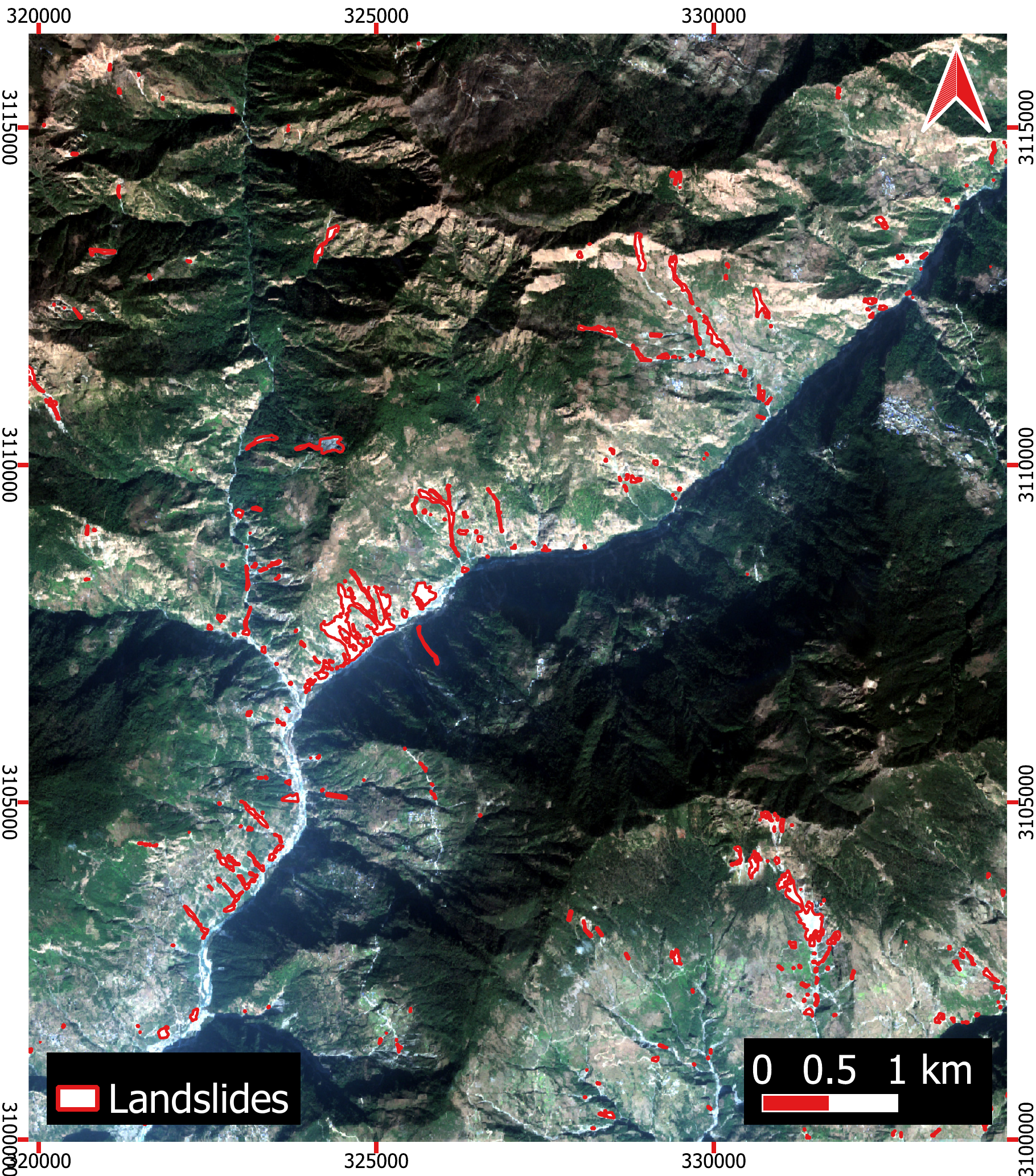}
\caption{Rasuwa District of Bagmati.}
\label{Fig: Rasuwa District of Bagmati}
\end{figure}

\subsubsection{Rasuwa District of Bagmati}
This study area is located in northern Kathmandu, Nepal, and is one of the most landslide-prone areas in the higher Himalayas. Most of the landslides in this region occurred in 2015 due to the Gorkha earthquake on April 25 and the Dolakha earthquake on May 12, with magnitudes of 7.8 and 7.3, respectively. These earthquakes, which also struck some northern parts of India, caused widespread landslides and about 8622 deaths in Nepal and 71 in India \cite{sharma2018mapping}. The landslides based on these two major earthquakes and their aftershocks were distributed mainly in the higher Himalayas, and our selected study area is one of the maximum affected areas. The study area is located in the Langtang National Park, with more than 31\% of forest land cover and several villages. Orographic monsoon precipitation plays a vital role in the climate of northern Kathmandu, with its annual average rainfall of more than 690 mm. The inventory dataset for this study area is compiled GPS data from an extensive field survey performed in the summer of 2018 and visual interpretation of some VHR images like RapidEye \cite{Ghorbanzadeh2019a}. 

\begin{figure}
\includegraphics[width=0.5\textwidth]{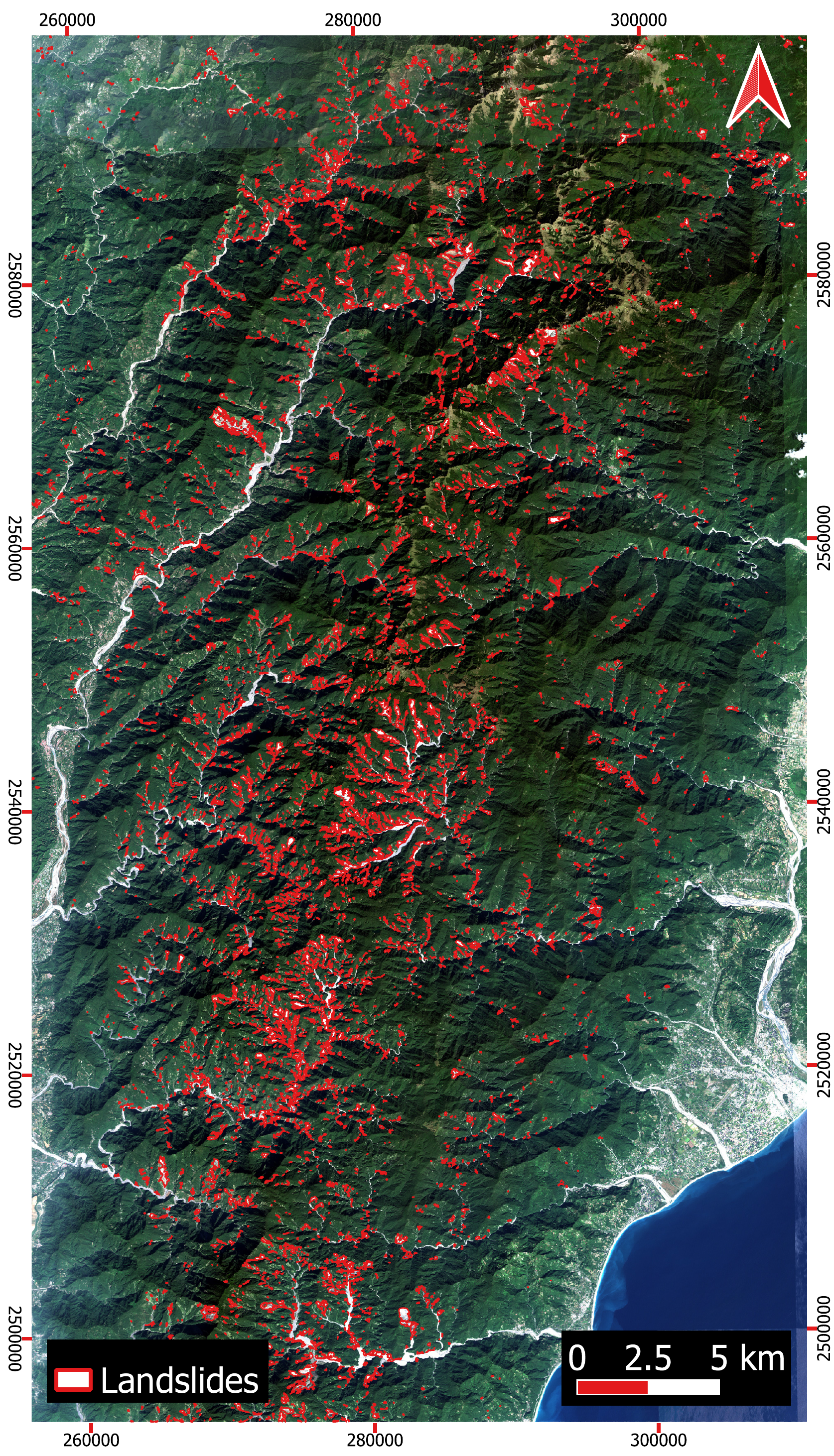}
\caption{Western Taitung County.}
\label{Fig: Western Taitung County}
\end{figure}

\subsubsection{Western Taitung County}
The case study area of Western Taitung County in Taiwan Province, like most of the other areas in this province, is heavily exposed to landslides caused mainly by typhoons and earthquakes. The annual mean precipitation of this subtropical climate island is 2500 mm, which is mainly based on summer and winter monsoons; the rainiest place receives 6,700 mm \cite{lin2011landslides, Ghorbanzadeh2021}. However, Typhoon Morakot, which occurred on August 7, 2009, brought over 2884 mm of precipitation to Western Taitung county and nearby Counties, mainly in southern areas, in only five days. Due to this intensive and concentrated precipitation, many serious landslides and floods occurred, causing severe damage to property and lives in this area. For instance, one of the enormous landslides covered the vilage of Hsiaolin, claiming about 400 lives. More than 50 people disappeared, and over 100 houses were destroyed entirely \cite{lin2011landslides}. Shallow-sliding landslides were the most frequent type of small landslides, while the deep-seated ones were mainly the larger ones triggered by the typhoon. Some studies have evaluated supervised and unsupervised DL solutions for detecting the landslides triggered by this typhoon in some parts of Western Taitung County (see, for example, Ghorbanzadeh \emph{et al.}  \cite{Ghorbanzadeh2021} and Shahabi \emph{et al.}  \cite{Shahabi2021}). Our landslide inventory dataset for this study area is compiled from previous studies and visual interpretation of Google Earth’s archive images from 2011 to 2013.

\subsection{Sensor Characteristics}
During the past three decades, different multi-spectral EO satellites missions like  Landsat 1--8 $\sim$ 1972, SPOT 1--7 $\sim$ 1986, and Ikonos 1999/2015 secured the required optical images for numerous remote sensing applications, such as climate change and land management \cite{Lissak2020}. Sentinel-2A, launched on 23 June 2015, is a European multi-spectral imaging mission that assures the continuity of image archives of the satellites mentioned above. Besides land monitoring and security services like global crop monitoring and land border surveillance, respectively, Sentinel-2 provides emergency management services for natural disasters such as fires, floods, and landslides. The twin-satellite capability of this satellite produces a high global revisit time of 5 days and 2--3 days at mid-latitudes if clouds do not cover the area. The images, acquired by multi-spectral instrument sensors, are provided in three different pixel spacings of 10, 20, and 60 m, within 13 bands. The spectral range is from visible bands starting from ultra blue with a central wavelength (CW) of 443 nm to near-infrared, and the short-wave infrared electromagnet spectrum by CW of 2190 nm in the 13's band \cite{Drusch2012}. The comprehensive spectral range, along with the high global revisit time, delivers the prime step in monitoring a wide range of remote sensing applications, and landslide detection in particular. Some free computing online platforms like Sentinel Hub (\href{https://www.sentinel-hub.com/}{https://www.sentinel-hub.com}) and Google Earth Engine (GEE) (\href{https://earthengine.google.com/}{https://earthengine.google.com}) provide users with Sentinel-2 images. These platforms also make it possible to find cloud-free images,  define the area of interest, and select and visualize different bands for small to medium workloads (Gomes et al., 2020). The GEE cloud system is an extensive multi-petabyte repository of remote sensing and geospatial public data, which is used in this study for acquiring the cloud-free Sentinel-2 images from our selected case study areas and different time series.

Advanced Land Observing Satellite (ALOS) was launched on January 24, 2006 by the JAXA. The acquired data is provided for different purposes such as land cover and forest monitoring, resource surveying, and hazard mapping, and uses three sensors. The phased array type L-band synthetic aperture radar (PALSAR) sensor supplies topographic data of DEM with a spatial resolution of 12.5 m. According to our literature review, the DEM and slope are the topographical information that is most widely applied for landslide mapping in combination with optical satellite imageries. Therefore, this study has generated the slope layer from the ALOS PALSAR (\href{https://search.asf.alaska.edu/#/}{https://search.asf.alaska.edu}). We have converted all considered optical and topographical layers to a 10 m pixel spacing.
\begin{figure*}[!htb]
\centering
\includegraphics[width=1\textwidth]{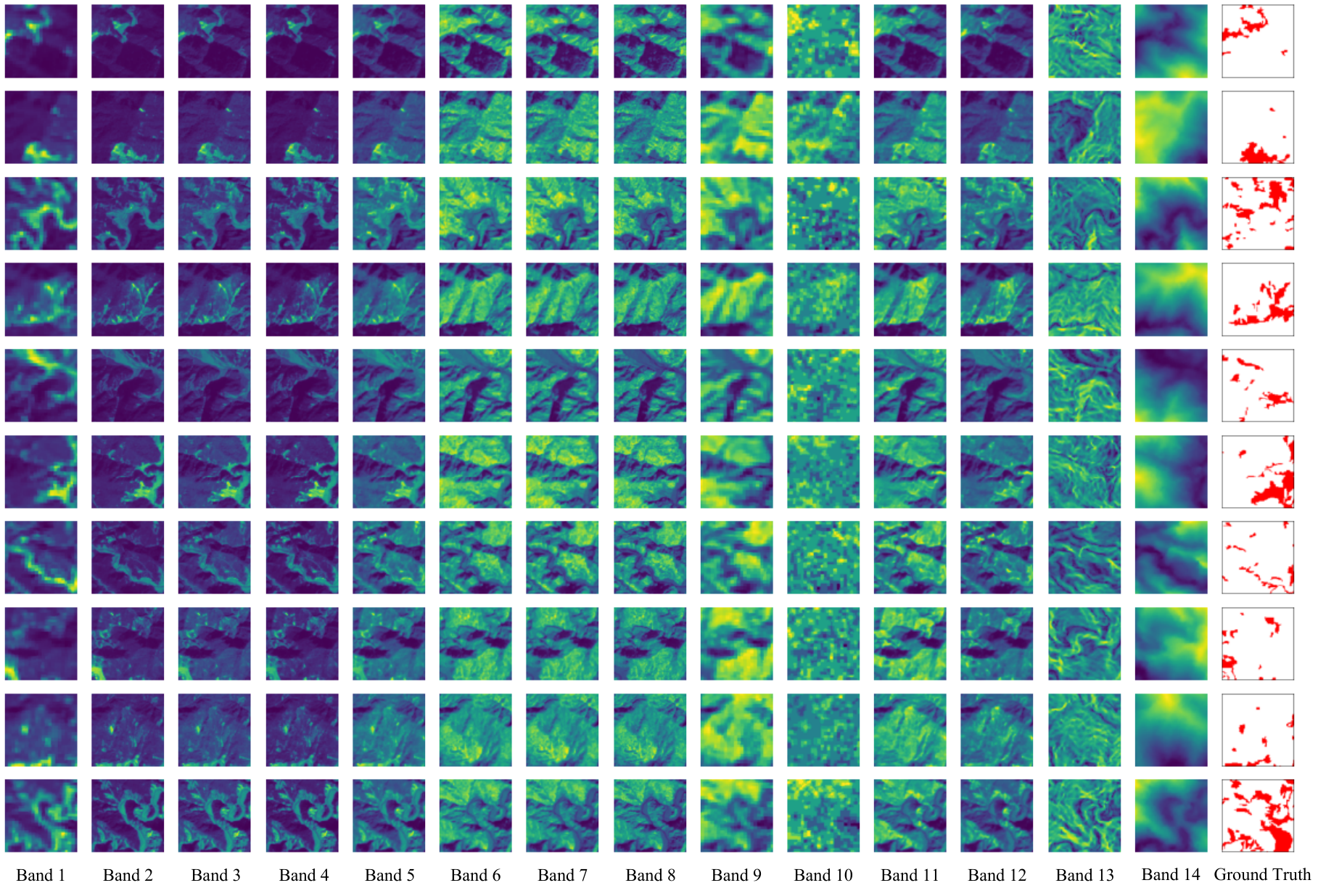}\\
\caption{Illustration of each single layer in the 128 $\times$ 128 window size patches of the collected landslide dataset. Bands 1-12 belong to the multi-spectral data from Sentinel-2 and bands 13-14 are slope and DEM data from ALOS PALSAR. The patches in the last column are corresponding labels and red polygons referring to the landslide category.}
\label{fig:layer}
\end{figure*}
\subsection{Landslide inventory annotation}
Developing an accurate landslide inventory dataset is the first step to implementing supervised DL models, and will facilitate the training and testing phases. Different approaches have been applied for this task by researchers in this field, such as an SVM \cite{mondini2017measures}, an OBIA framework \cite{Shi2020}, a data preprocessing workflow \cite{Yi2020a}, and manual interpretation \cite{Catani2021, ghorbanzadeh2019evaluation,Lei2019, Prakash2021, Qi2020,Ullo2021}. Manual interpretation based on satellite imagery is the most common approach in similar studies. However,  the inventory can be changed easily by the preferences of different landslide experts \cite{tanyacs2019factors}. To avoid this issue, we followed a two-step workflow for landslide annotation from satellite imagery. We first designed and applied an OBIA framework for landslide annotation, and then manually verified and corrected the resulting landslide polygons one by one for all study areas. Our first step in OBIA is to calculate some relevant image difference indices based on pre- and post-landslide images for each study area. The resulting indices and post-landslide images were then used for image segmentation using the multiresolution segmentation method. The landslide bodies were extracted using a rule-based image classification to compute different thresholds for each study area. A similar OBIA framework has been done by \cite{Ghorbanzadeh2022a} for landslide detection from bio-temporal images. In the manual interpretation step, other data sources available for each study area, such as Google Earth images and the landslide inventory geodatabase generated by \cite{Zhang2019c} were then used to visually correct landslide polygons. Thus, our annotations refer to the polygons showing the exact location and boundaries of landslides and lack any other information, such as forming material, landslide type, or the volume of mass movement.  
\begin{figure*}[!htb]
  \centering
  \includegraphics[width=\linewidth]{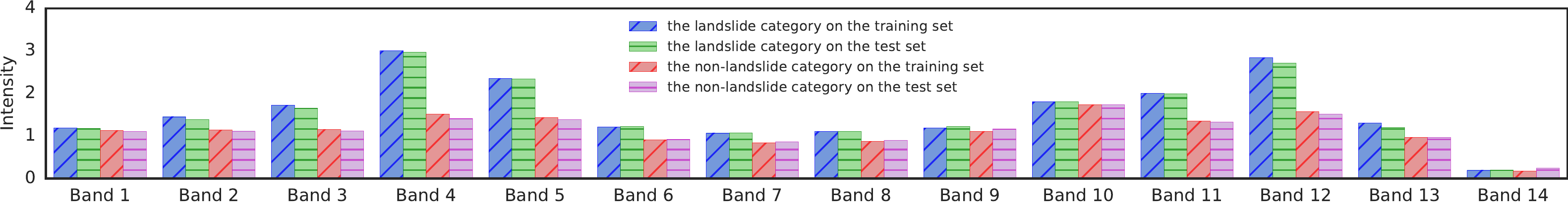}\\
  (a) Iburi-Tobu area of Hokkaido\\
  \includegraphics[width=\linewidth]{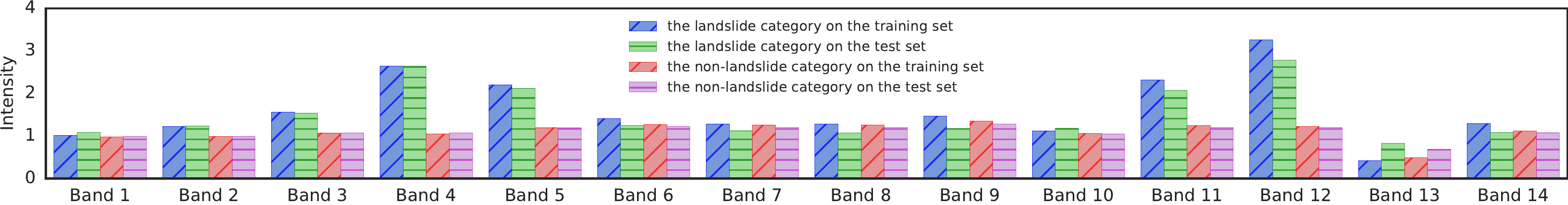}\\
  (b) Kodagu District of Karnataka\\
  \includegraphics[width=\linewidth]{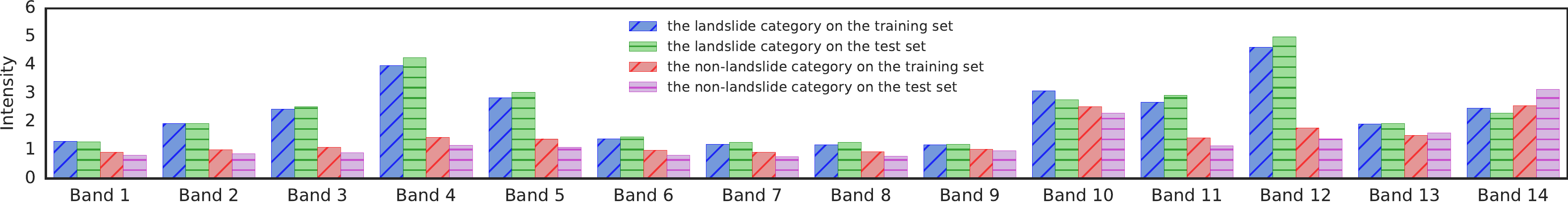}\\
  (c) Rasuwa District of Bagmati\\
  \includegraphics[width=\linewidth]{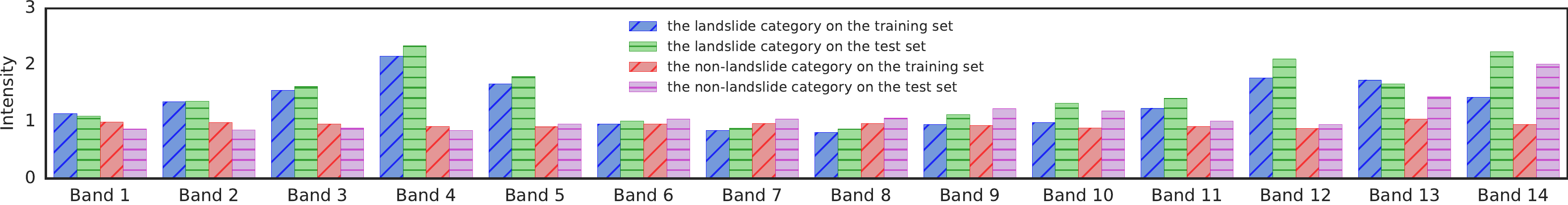}\\
  (d) Western Taitung County\\
  \includegraphics[width=\linewidth]{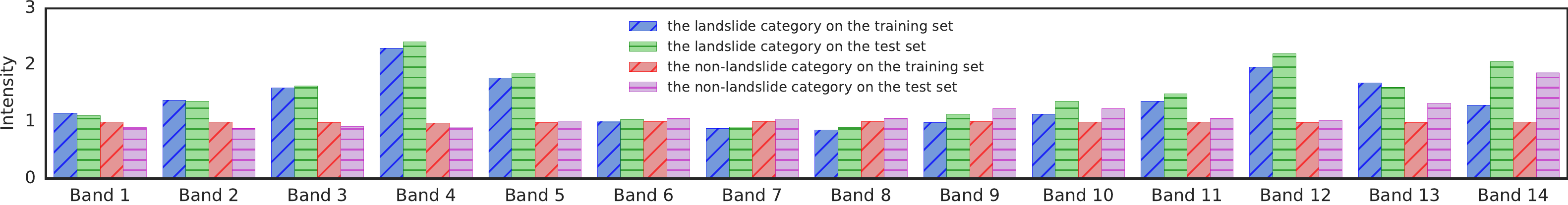}\\
  (e) All study areas
  \caption{The mean value of each band for the landslide samples and the non-landslide samples in each study area and all study areas. (a) Iburi-Tobu. (b) Kodagu. (c) Rasuwa. (d) Taitung. (e) All study areas.}
\label{fig:mean_value}
\end{figure*}
\subsection{Benchmark Dataset Statistics and Shape}
The main goal of this paper is to provide a large, diverse dataset of annotated landslide images that can be used as a benchmark dataset to train ML and DL models for the landslide detection task. As we looked for freely available satellite imagery, we selected Sentinel-2 as the satellite imagery sharing platform with the greatest worldwide availability and finest spatial resolution. To build a multi-source benchmark dataset, we collect not only optical Sentinel-2 imagery but also the necessary topographical information from the digital elevation model (DEM) and slope layer derived from ALOS PALSAR. In doing so, some challenging issues, including different spatial resolutions, and range of pixel values, are investigated. As a result, the pixel spacing of the topographical layers of DEM and slope are converted to 10 m, similar to that used for all of the Sentinel-2 bands. Thus, we could include different case study areas from different geographical regions in the world with a diversity of landslide triggers, topography, and land cover, which is more challenging than working on one landslide-affected case study area. In total, 3799 annotated patches are provided without considering any overlap. We use 959 patches to train selected state-of-the-art DL segmentation models, while the remaining 2840 patches are used for evaluation in the experiments. The training patches are the first quarter generated patches from each case study area.
Each image patch is a composite of 14 bands that include: multi-spectral data from Sentinel-2 (band 1 to band 12) as well as slope and DEM data from ALOS PALSAR (bands 13 and 14). As already mentioned, all bands in the dataset are resized to the resolution of 10 meters per pixel. The image patches have the size of 128 $\times$ 128 pixels and are labeled pixel-wise. Ten sample patches are selected from different study areas to represent these 14 bands and the corresponding label in Figure \ref{fig:layer}. 
\begin{figure*}
  \centering
  \includegraphics[width=\linewidth]{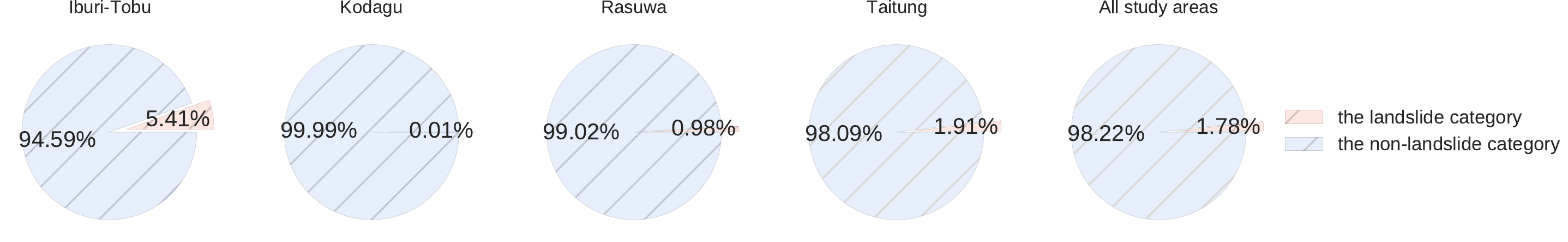}\\
  (a) Training set\\
  \vspace{2em}\includegraphics[width=\linewidth]{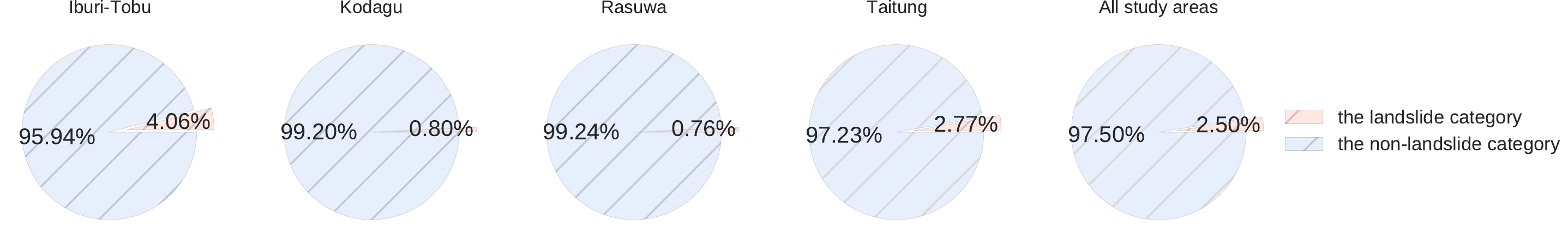}\\
  (b) Test set
  \caption{The ratio of the landslide samples to the non-landslide samples for each study area and the average for all study areas. (a) Training set. (b) Test set.}
\label{fig:ratio}
\end{figure*}
The landslide/non-landslide split of the mean pixel value intensity for each band can be found in Figure \ref{fig:mean_value}, which exhibits the distribution of training and testing sets of each single study area and all together for the whole dataset. The blue and green bars represent the pixel value intensity of the landslide category in the training and test samples, respectively, and the red and pink bars denote those of the non-landslide category. A landslide often removes the surface land cover and vegetation along the mass movement path and leaves specific signatures.  The affected areas are mostly brighter than the non-landslide background areas as a result of the loss of surface ground and vegetation and the exposed sliding surface that contains the fresh soil and rock from beneath  \cite{Tehrani2021}. 
The mean pixel value of the landslide category in both training and testing sets is higher than that of the non-landslide category in almost all bands. Thus, the landslide features show higher intensity compared to that of non-landslide background areas. This difference varies for different bands; the highest difference appears in band 4 and band 5.  Sentinel-2 sensors contain three bands in the ``red edge'' wavelength range--band 5, band 6, and band 7--which are important for vegetation-related studies and agriculture applications. Figure \ref{fig:mean_value} shows that the first red edge band of band 5 has a higher difference between the considered categories than band 6 and band 7.
Moreover, the short-wave infrared wavelength from 1375 nm to 2190 nm (band 10, band 11, and band 12) represents a high mean pixel value for both categories with a high difference in band 12, which is more obvious in the Rasuwa District of Bagmati. Therefore, visible and near-infrared bands (band 6, band 7, and band 8) illustrate the lowest difference between landslide and non-landslide; these features share almost the same mean pixel values. 
As mentioned above, band 13 and band 14 refer to the slope and elevation data. We have covered a large breadth of diversity in elevation data in Landslide4Sense, which is apparent in the differences among the different study areas and the Iburi-Tobu area of Hokkaido and Western Taitung County in particular. This diversity is also noticeable in the differences between the elevations of landslides incurring in various study areas. In the case of the Kodagu District of Karnataka, there is no huge difference between the elevation of landslide and non-landslide features. While in the Rasuwa District of Bagmati landslides usually happen in lower areas along the rivers. More information about the topography of each study area can be found in section A. Study areas. The way of splitting the areas for training and testing datasets also has increased the difference between the elevation data in these datasets and this matter is evident in the case of Western Taitung County.
The diversity of the prepared landslide benchmark dataset is not limited to criteria like triggers, geographic location, and topographic characteristics, but also the shape, size, distribution, and frequency of the landslides appearing in each study area (see study areas in Section A). Figure \ref{fig:ratio} shows the ratio of the landslide category to the non-landslide one. As mentioned above, one-quarter of each study area is dedicated to the training dataset and three-quarters is preserved for testing the models. Thus,  Figure 8 also illustrates the ratio for both the training and the testing sets. Iburi-Tobu area of Hokkaido has the highest landslide to non-landslide ratio and close to 5.5\% of the training section of this study area is covered by landslides, which is apparent from Figure \ref{Fig: Iburi-Tobu area of Hokkaido}. In contrast, only a very small amount (0.01\%) of the training section of the Kodagu District of Karnataka is placed in the landslide category. This situation is almost the same in the test set: the Iburi-Tobu area of Hokkaido and Kodagu District of Karnataka have the highest and the lowest number of landslide areas, respectively. We purposely split the training and testing sets into one-quarter and three-quarters to make the task more challenging. However, using other splitting techniques such as random selection or using k-fold cross validation would increase the similarity among training and test sets and make the procedure safer.  Meanwhile, we expect the whole dataset (both training and test sets) to be applied to training the desired models for novel purposes with or without using a local labeled training set. The landslide category is represented by red shading and the non-landslide category is shown by blue color in Figure \ref{fig:ratio}.

\section{Methodology}
Recent advances in computer vision and machine learning have facilitated rapid progress in the landslide detection task. To provide a preliminary analysis of how advanced deep learning methods can be applied to the benchmark dataset collected in this study, we choose ten representative deep learning methods for experimentation. In this section, we will briefly introduce each of these advanced deep learning approaches.

\subsection{FCN-8s}
FCN-8s is the first fully convolutional network (FCN) designed for the semantic segmentation task \cite{long2015fully}. The key idea is to create a convolutional network, which is able to receive an input image of an arbitrary size and produce pixel-to-pixel predictions. To balance the learning between shallow and deep features and produce more accurate and detailed segmentation results, FCN-8s proposes a skip architecture that combines coarse and fine layers of semantic information. 

\subsection{PSPNet}
The pyramid scene parsing network (PSPNet) aims to address the issue of scene parsing with unrestricted open vocabulary and variety of scenes \cite{zhao2017pyramid}. Using the pyramid pooling technique, PSPNet aims at learning global context information using a regional aggregation method. With this technique, PSPNet can fuse features under four different pyramid scales, achieving the purpose of multi-scale feature learning.

\subsection{ContextNet}
The ContextNet can produce competitive semantic segmentation in real-time with low memory usage using factorized convolution, network compression, and pyramid representation \cite{poudel2018contextnet}. The main idea of ContextNet is to combine a deep network at the low resolution  that obtains relevant context effectively with a shallow network at high resolution that focuses on semantic details. In this way, ContextNet can perceive context at low resolution and refine it for high-resolution results.

\subsection{DeepLab-v2}
DeepLab-v2 is a powerful model for semantic segmentation \cite{chen2017deeplab}. The main components of this method include atrous convolution, the atrous spatial pyramid pooling (ASPP) module, and a fully connected conditional random field (CRF). With the help of atrous convolution, DeepLab-v2 has a special ability to reduce the number of parameters and computation required while extending the field of view of filters. Then, the ASPP module probes the convolutional layer with filters at different sampling rates, thus capturing multi-scale features. Finally, the fully connected CRF is adopted to further refine the segmentation performance.

\subsection{DeepLab-v3+}
To capture sharper object boundaries, the DeepLab-v3+ model \cite{chen2018encoder} adopts a novel encoder-decoder with atrous separable convolution. Specifically, With the encoder module, the feature maps are reduced in size and high-level semantic information is captured, while the decoder module eventually fills in the spatial information. Additionally, depth-wise separable convolution is applied to increase computational efficiency. 

\subsection{LinkNet}
LinkNet is an efficient semantic segmentation model designed for real-time visual scene understanding \cite{chaurasia2017linknet}. The main idea of LinkNet is to directly bypass spatial information from the encoder to the corresponding level of the decoder to improve the interpretation accuracy while decreasing the processing time. With the proposed bypassing technique, LinkNet can effectively maintain the object boundaries in the image without additional training parameters.

\subsection{FRRN}
Full-resolution residual network (FRRN) is a novel ResNet-like method for semantic segmentation \cite{pohlen2017full}. The main idea of FRRN is defined by the purpose of combining multi-scale context with pixel-level information by a two-stream architecture. Specifically, the first stream aims to learn precise boundary information at the full image resolution, while the second stream consists of a series of pooling layers to extract high-level semantic features for recognition. In the prediction phase, these two streams are fused at the full image resolution by residual learning. 

\subsection{SQNet}
SQNet is a light-weighted semantic segmentation architecture for autonomous driving with high efficiency \cite{sqnet}. The main components of SQNet include the ELU activation functions, a SqueezeNet-like encoder, parallel dilated convolutions, and a decoder with SharpMask-like refinement modules. Specifically, the parallel dilated convolution layer combines the feature maps at different receptive field sizes by applying four dilated convolutions with distinct dilation factors. In this way, SQNet can achieve multi-scale feature learning with relatively light-weighted parameters.

\subsection{U-Net}
U-Net is an FCN that was initially designed by Ronneberger \emph{et al.} \cite{ronneberger2015u} for biomedical image segmentation and now is more commonly used for a wide range of remote sensing image classification and object detection tasks such as landslide detection. The U-Net architecture contains an encoder and a decoder path. The former is a contracting path similar to the typical architecture of a CNN and responsible for capturing low-level representations. In contrast, the second half of the architecture is an expanding path for capturing high-level representations. Like other FCNs, the U-Net architecture is strengthened by skip connections among the encoder and decoder paths to fine-grained aggregate details from the encoder path to the corresponding layers in the decoder path.

\subsection{ResU-Net}
The ResU-Net combines Residual Networks (ResNets) \cite{He2016} and U-Net. This variant of the U-Net features structures based on residual learning blocks that are replaced with the plain convolution layers \cite{zhang2018road}. The skip connection in these blocks does not add any other parameter to the next layer except the output obtained from the previous one. The purpose of this combination is to enhance learning capabilities, which can also prevent gradient vanishing. The ResU-Net has garnered much interest in remote sensing communities for landslide detection from satellite imagery \cite{Qi2020, Ghorbanzadeh2021}.

\section{Experiments}
\subsection{Experimental Design}

\begin{figure*}
  \centering
  \includegraphics[width=0.95\linewidth]{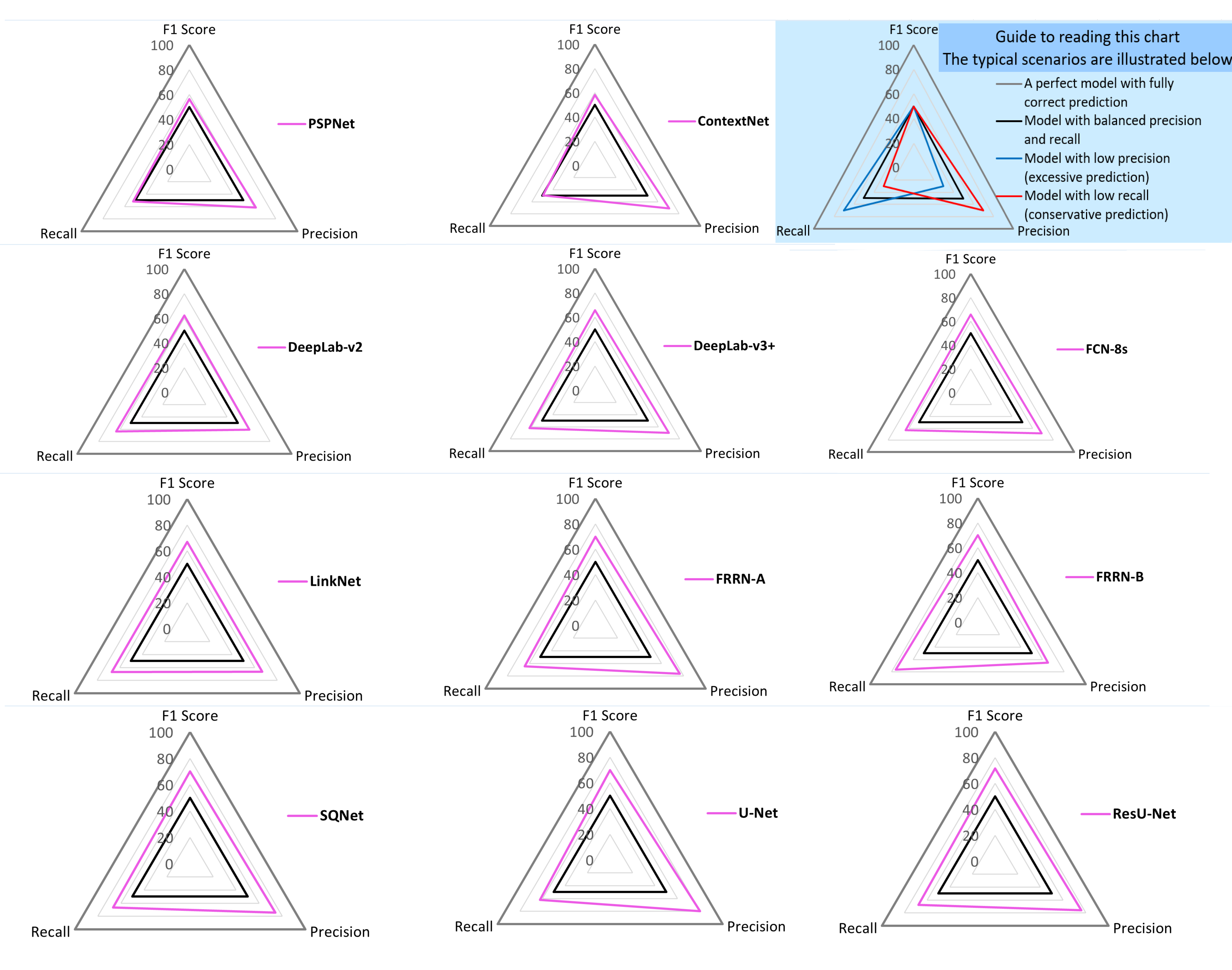}\\
  \caption{Spider plots representing the performance of the applied DL models. The precision and recall scores are plotted on the X-axis (left and right angle of three angles), and the F1 scores is plotted on the Y-axis (top angle). The blue box explains the typical scenarios.}
\label{fig:Accuracy}
\end{figure*}

The experiments of this study were carried out in the context of introducing the Landslide4Sense benchmark dataset for landslide detection by comparing the performance of the state-of-the-art DL segmentation models. To achieve this multi-source dataset, we stacked optical data of 12 bands from Sentinel-2 and also the topographic data of two layers of slope and DEM from the ALOS PALSAR into one volume in order to create image patches. To this end, all bands are resized to the resolution of 10 meters per pixel, and a dataset Comprising 14 bands is prepared. In the experiments, all these 14 bands were used for training the DL models as well as for the evaluation procedure. We trained 11 DL models on the training set and tested them all based on the hold out sample patches. Unlike previous studies \cite{Ghorbanzadeh2021, Prakash2021} we did not perform any transferability evaluation, as here we intend to introduce Landslide4Sense benchmark dataset for training the state-of-the-art DL segmentation models. Therefore, we trained the DL models on a combined labeled training set of study areas from four different geographical regions. In general, the applied DL models were able to learn the features expected to discriminate landslides from the surrounding areas in the test areas. The Adam optimizer by considering a learning rate of $1e-3$ and a batch size of 32 was used for all of the applied models to compare them fairly. We applied training from scratch for all models with 5000 iterations. 
All experiments in this study were conducted on a cluster with four NVIDIA Tesla V100 GPUs. The results of the landslide detection performance of the applied DL models were provided in terms of the three most standard accuracy assessment metrics: a) precision, b) recall, and c) F1-score, all of which have been used in several landslide detection studies.

\begin{table}
\centering
\caption{Quantitative results of different deep neural networks for the landslide detection (\%).}
\label{tab:result}
\setlength{\tabcolsep}{5.5mm}{
\begin{tabular}{cccc}
\hline
 & Recall & Precision & $F_1$ \\ \hline
PSPNet \cite{zhao2017pyramid} & 52.03 & 61.55 & 56.39 \\ 
ContextNet \cite{poudel2018contextnet} & 49.29 & 70.77 & 58.11 \\ 
DeepLab-v2 \cite{chen2017deeplab} & 63.68 & 60.8 & 62.21 \\ 
DeepLab-v3+ \cite{chen2018encoder} & 62.11 & 69.91 & 65.78 \\ 
FCN-8s \cite{long2015fully} & 63.05 & 68.66 & 65.73 \\ 
LinkNet \cite{chaurasia2017linknet} & 67.02 & 66.76 & 66.89 \\ 
FRRN-A \cite{pohlen2017full} & 64.4 & 76.57 & 69.96 \\ 
FRRN-B \cite{pohlen2017full} & 76.16 & 64.93 & 70.1 \\ 
SQNet \cite{sqnet} & 66.69 & 74.2 & 70.24 \\
U-Net \cite{ronneberger2015u} & 62.17 & 79.91 & 69.94 \\
ResU-Net \cite{zhang2018road} & 67.71 & 76.08 & 71.65 \\ \hline
\end{tabular}%
}
\end{table}

\begin{figure*} 
  \centering
  \includegraphics[width=0.95\linewidth]{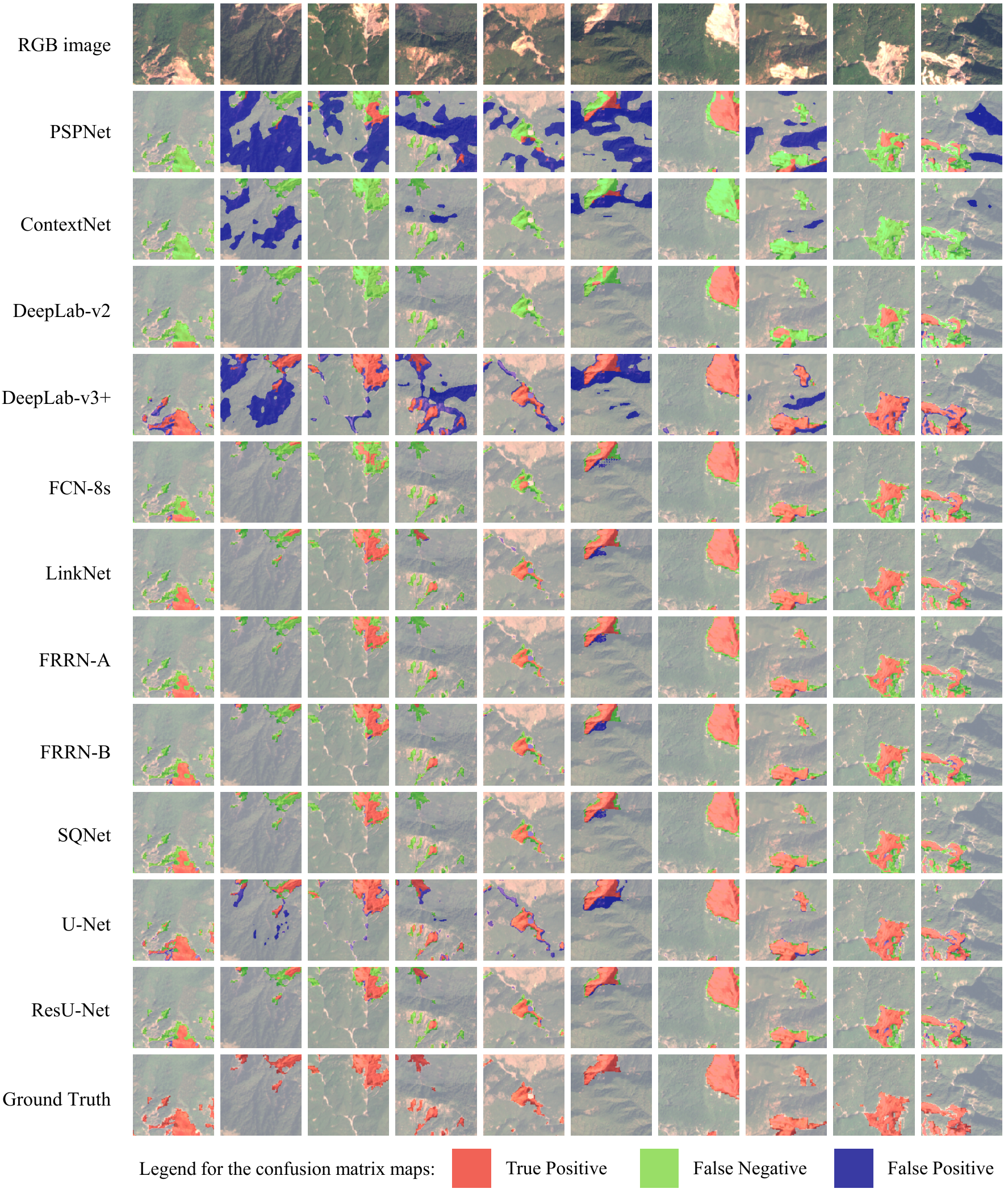}\\
  \caption{The landslide detection maps obtained by different deep learning methods.}
\label{fig:maps}
\end{figure*}

\subsection{Experimental Results}
Quantitative results of landslide detection by the applied DL models are reported in Table \ref{tab:result}.  As can be seen, ResU-Net accomplishes the best performance, with an F1-score of 71.65\%, in the landslide detection tasks followed by the SQNet and its F1-score of 70.24\%, which marginally surpasses the third-best model, FRRN-B, by 0.14 percentage points. For the accuracy assessment metric of the precision, U-Net shows a stupendously superior performance with precision of almost 80\%, which is more than 3 percentage points greater than the second-highest precision, obtained by the FRRN-A. Precision in terms of recall is not as good as for precision and the resulting recall values from most of the models are noticeably lower than those of precision. This difference means that although the pixels detected as landslides can be reliable with high confidence, the models were not able to identify many other labeled landslide pixels. The highest difference can be seen in the case of ContextNet, with its precision and recall values of 70.77\% and 49.29\%, respectively. Note that many DL methods tend to yield good performance on only one of these two metrics while sacrificing the accuracy of the other metric. Take the FRRN-A model for example. While it obtains a precision of 76.57\%, its recall is only 64.40\%, leading to an F1 score of only 69.96\%. By contrast, ResU-Net can achieve relatively more balanced results on both recall and precision metrics, leading to the highest F1 score. Thus, comparing the precision and recall values can tell us more about whether the model has an excessive or a conservative prediction.
Figure \ref{fig:Accuracy} visualizes these results for all applied DL models. The pink triangles in this Figure represent each model’s values for the applied accuracy assessment metrics. 
We further visualize the prediction of all models based on some randomly selected landslides including patches from the whole test set in Figure \ref{fig:maps}. The labels (true positives), landslide pixels that models are not able to detect (false negative), and non-landslide pixels that are detected as landslides incorrectly (false positive) by the models are shown in red, green, and blue, respectively. It can be observed from Figure 10 that making a good balance between the precision and recall is very difficult for most of the DL methods used in this study. Take the landslide detection maps of DeepLab-v3+ as an example. Although DeepLab-v3+ can detect most landslide regions well, it tends to yield a high false-alarm rate, and many non-landslide pixels are misclassified as the landslide category (colored in blue). By contrast, ResU-Net can well suppress false positive predictions while preserving a high accuracy in detecting landslide regions. It can be observed from the 12th row in Figure 10 that there are relatively fewer pixels colored in green or blue than in detection maps of other methods. This phenomenon also demonstrates that ResU-Net can better balance precision and recall, achieving the best detection performance on the Landslide4Sense dataset.

\subsection{Comparison with Related Work}
Since the Sentinel-2 data that is used in Landslide4Sense is available for free, it ensures easy adaptability and facilitates automation of any trained model based on this dataset for landslide detection during cloud-free days in novel regions without being dependent on high-resolution and VHR data like UAV and commercial satellite imagery. However, to date a limited number of studies using Sentinel-2 data for landslide detection have been conducted due to its medium spatial resolution, which is considered one of the main factors in reducing the detection performance \cite{Prakash2021}. Therefore, detecting landslides from Sentinel-2 data is more challenging than doing so with high-resolution satellite imagery like Rapid-eye \cite{Soares2020}. Prakash \emph{et al.} \cite{prakash2020mapping} used Sentinel-2 images along with a high-resolution Lidar DEM for landslide detection in a heavily forested area in Douglas County, Oregon, USA. Semi-automated pixel and object-based methods were compared with a modified U-Net and the resulting highest F1-score values were 51.3\%, 54.6\%, and 56.2\%. Their applied DL method obtained a higher F1 score than the conventional methods, which is close to our lowest classification result of almost 56.4 \% obtained by PSPNet. A comprehensive transferability evaluation of two FCN models (U-Net and ResU‑Net) has been done in \cite{Ghorbanzadeh2021} for landslide detection using Sentinel‑2 data. In this study, the effectiveness of the applied models was compared based on different training and test sets selected from three different geographical areas. The highest F1‑score value of 73.32\% was achieved by ResU‑Net, which was trained and tested on two separate datasets. This model also has obtained the highest F1‑score value among the 11 trained DL models in our study. However, the overall mean of the F1-score values for all performed scenarios ranges from 56.6\% to 70\% and from 61.36\% to 71.2\% for U‑Net and ResU‑Net, respectively. The higher mean F1-score values of the ResU‑Net over those of the U‑Net in this work also confirm our resulting F1-score values based on these two models.

\section{Discussion and Conclusions}
\begin{figure}
  \centering
  \includegraphics[width=0.45\textwidth]{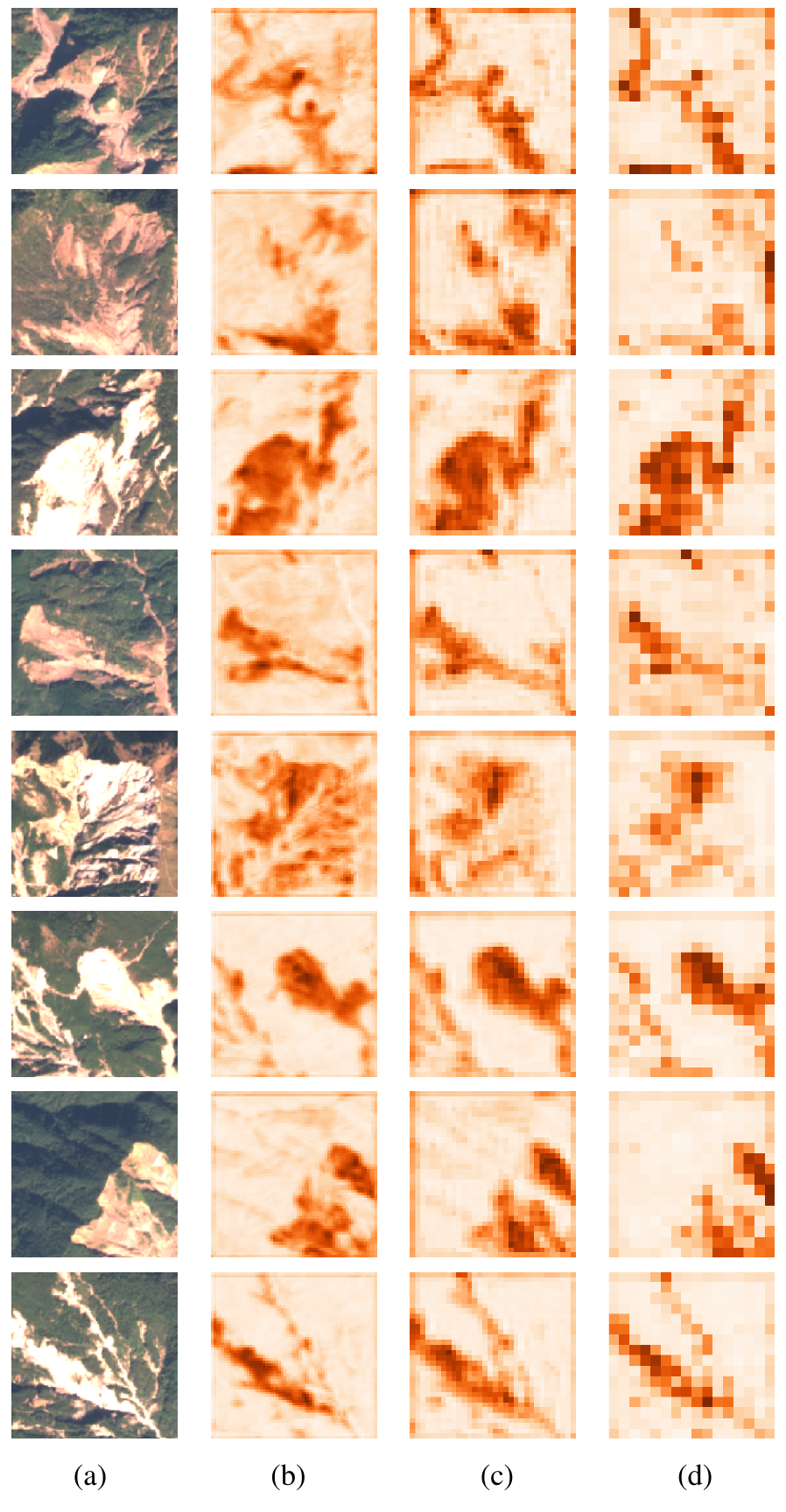}\\
  \caption{Visualization maps of the convolutional features at different layers. We here explore the PSPNet as an example. (a) RGB image. (b) Features of the first residual block. (c) Features of the second residual block. (d) Features of the third residual block.}
\label{fig:feature}
\end{figure}

We here introduced the novel reference benchmark Landslide4Sense, featuring 3,799 Sentinel-2 and ALOS PALSAR image patches from various geographical regions worldwide. Eleven state-of-the-art DL segmentation models were used to demonstrate the benchmark for landslide detection: PSPNet, ContextNet, DeepLab-v2, DeepLab-v3+, FCN-8s, LinkNet, FRRN-A, FRRN-B, SQNet, U-Net, and ResU-Net. To better understand the information extracted by the deep learning models in their different layers, we further visualized the convolutional features in Figure \ref{fig:feature}, exploring the feature maps of PSPNet as an example. This nicely showed how in the shallow layer the first residual block extracted features with more detailed spatial information about potential landslide areas, especially in the boundary regions. By contrast, the features from deeper layers like the third residual block seem to be more abstract. The detailed boundary of the landslide regions can hardly be maintained as the spatial size of the feature maps decreases significantly in the deeper layers. 
This phenomenon also demonstrates the importance of combining both shallow and deep features when addressing the landslide detection task. The quantitative results in this study also verify this finding. For example, both U-Net and ResU-Net show superior landslide detection results compared to other state-of-the-art methods owing to the well-designed `U-shape' architecture, which helps to balance the information of both shallow layers and deep layers.

Deep Learning models have become ubiquitous in remote sensing image segmentation and classification tasks like landslide detection. However, DL models need much training data, and the lack of publicly available labeled data and annotated images is usually a big barrier to using these models. Increasing the availability of annotated landslide images is thus fundamental to obtaining acceptable results from DL models in landslide detection. We here provide a unique benchmark dataset that can continuously be expanded and updated, opening new avenues for research in the landslide community. The landslide benchmark dataset represents a significant advancement for the use of DL models in landslide detection. It supplies a promising data source to support research studies in the field of landslide detection and hazard assessment. The landslide benchmark dataset is suitable for 1) training DL models and using them for novel unexplored geographical regions in emergency cases (transfer learning); 2) easy extension to larger scales by continuous addition of a wide range of image patches from landslide affected areas, incorporating further data from the publicly available Sentinel-2 and ALOS PALSAR images; and 3) providing an explicit norm to compare the generalization abilities of new DL models proposed for remote sensing image segmentation and classification, including unsupervised, self-supervised, and semi-supervised methods for landslide detection.

It must be emphasized that since we provide single-shot landslide images, Landslide4Sense has limitations for change detection applications that require multi-temporal images to accurately annotate changes that happened on the ground surface, such as landslides. Adding additional information to the data set, such as forming material in input data and the type of landslide in labels, may result in slightly different predictions. We would also point out that finding cloud-free Sentinel-2 images forced us to acquire images from times different to the exact time of a landslide event. In addition, data collectors need to provide clear guidance for users about the domain where the trained models based on the dataset are expected to lead to effective detection results. Therefore, as a final point, we emphasize that although our selected regions affected by landslides differ in various aspects such as topography, triggers, and landslide shape and size, most of them have high vegetation coverage, which is the main reason for the often-limited model transferability to novel areas without vegetation land cover. Then, the transferability of models trained on this dataset alone remains a big challenge for novel regions that are less vegetated. This however, as mentioned above, is mitigated by the fact that the provided dataset lends itself to extension using additional examples from freely available Sentinel-2 images acquired for new landslide events.

\section*{Acknowledgment}
The authors would like to thank the Institute of Advanced Research in Artificial Intelligence (IARAI) for its support.

\bibliographystyle{IEEEtran}
\bibliography{landslide4sense}

\end{document}